\documentclass[10pt,twocolumn,letterpaper]{article}

\usepackage[pagenumbers]{cvpr} %

\usepackage{algorithm}
\usepackage{algorithmic}
\usepackage{url}            %
\usepackage{booktabs}       %
\usepackage{amsfonts}       %
\usepackage{amssymb}     
\usepackage{amsmath}     
\usepackage{nicefrac}       %
\usepackage{microtype}      %
\usepackage{xcolor}         %
\usepackage{bm}     
\usepackage{xfrac} 
\usepackage{array}
\usepackage{makecell}
\usepackage{multirow}
\usepackage{enumitem}
\usepackage{color}
\usepackage{colortbl}

\definecolor{lightblue}{RGB}{122,46,246}
\definecolor{lightgreen}{RGB}{0,128,0}
\definecolor{lightred}{RGB}{226,3,9}
\definecolor{lightorange}{RGB}{225,165,0}
\definecolor{lightours}{RGB}{255,246,247}
\definecolor{FLOPs}{RGB}{248,203,173}

\newcommand{\PreserveBackslash}[1]{\let\temp=\\#1\let\\=\temp}
\newcolumntype{C}[1]{>{\PreserveBackslash\centering}p{#1}}
\newcolumntype{R}[1]{>{\PreserveBackslash\raggedleft}p{#1}}
\newcolumntype{L}[1]{>{\PreserveBackslash\raggedright}p{#1}}

\definecolor{cvprblue}{rgb}{0.21,0.49,0.74}
\usepackage[pagebackref,breaklinks,colorlinks,allcolors=cvprblue]{hyperref}

\def\paperID{*****} %
\def\confName{CVPR}
\def\confYear{2025}

\title{Mixture of Ranks with Degradation-Aware Routing for One-Step Real-World Image Super-Resolution}

\author{
    Xiao He\textsuperscript{1}, Zhijun Tu\textsuperscript{2}, Kun Cheng\textsuperscript{1}, Mingrui Zhu\textsuperscript{1}, Jie Hu\textsuperscript{2} Nannan Wang\textsuperscript{1}\footnotemark[2], Xinbo Gao\textsuperscript{1}, \\
    \textsuperscript{1}State Key Laboratory of Integrated Services Networks, Xidian University \\
    \textsuperscript{2}Huawei Noah's Ark Lab \\
    \tt\small xiaohe366@gmail.com, zhijun.tu@huawei.com, nnwang@xidian.edu.cn
}

\begin{document}
\maketitle

\renewcommand{\thefootnote}{\fnsymbol{footnote}}
\footnotetext[2]{Corresponding author.}
\renewcommand{\thefootnote}{\arabic{footnote}}

\begin{abstract}
   The demonstrated success of sparsely-gated Mixture-of-Experts (MoE) architectures, exemplified by models such as DeepSeek and Grok, has motivated researchers to investigate their adaptation to diverse domains. In real-world image super-resolution (Real-ISR), existing approaches mainly rely on fine-tuning pre-trained diffusion models through Low-Rank Adaptation (LoRA) module to reconstruct high-resolution (HR) images. However, these dense Real-ISR models are limited in their ability to adaptively capture the heterogeneous characteristics of complex real-world degraded samples or enable knowledge sharing between inputs under equivalent computational budgets. To address this, we investigate the integration of sparse MoE into Real-ISR and propose a Mixture-of-Ranks (MoR) architecture for single-step image super-resolution. We introduce a fine-grained expert partitioning strategy that treats each rank in LoRA as an independent expert. This design enables flexible knowledge recombination while isolating fixed-position ranks as shared experts to preserve common-sense features and minimize routing redundancy. Furthermore, we develop a degradation estimation module leveraging CLIP embeddings and predefined positive-negative text pairs to compute relative degradation scores, dynamically guiding expert activation. To better accommodate varying sample complexities, we incorporate zero-expert slots and propose a degradation-aware load-balancing loss, which dynamically adjusts the number of active experts based on degradation severity, ensuring optimal computational resource allocation. Comprehensive experiments validate our framework's effectiveness and state-of-the-art performance.
\end{abstract}

\section{Introduction}

In image super-resolution (SR) tasks~\cite{,dong2014learning,zhang2015feature,wang2018esrgan,liang2021swinir}, models process low-resolution (LR) inputs to reconstruct high-resolution (HR) outputs with enhanced high-fidelity details. Traditional SR methodologies~\cite{kim2016accurate,bao2022attention,chen2023activating} typically generate LR images via bicubic downsampling of HR counterparts. However, real-world image degradation processes exhibit inherent complexity, encompassing multifaceted distortions such as blurring, sensor noise, and other ill-defined artifacts. This discrepancy between synthetic and real-world degradation has positioned real-world image super-resolution (Real-ISR)~\cite{wang2021real,zhang2021designing} as a critically challenging problem, driving substantial research efforts in recent years.

\begin{figure}
  \includegraphics[width=0.92\linewidth]{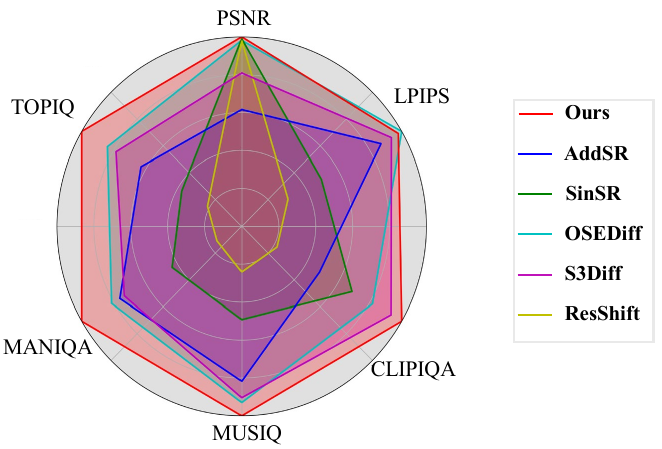}
  \caption{Performance Comparison. Compared to other Real-ISR methods, MoR-DASR achieves superior performance with just a single diffusion step.}
  \label{fig:teaser}
\end{figure}

Generative adversarial networks (GANs)~\cite{goodfellow2020generative} and diffusion models (DMs)~\cite{ho2020denoising,song2020denoising} currently constitute the predominant architectures for Real-ISR tasks, with diffusion models garnering particular research interest due to their superior generative priors. However, the iterative sampling process inherent to DMs—typically requiring multiple sequential steps—imposes prohibitive computational latency, hindering practical deployment. Recent efforts~\cite{wang2023sinsr,wu2024one,he2024one} mitigate this limitation through single-step SR frameworks, achieved either by distilling multi-step diffusion models or fine-tuning pre-trained diffusion models via Low-Rank Adaptation (LoRA) modules~\cite{hu2022lora}. While these advancements have accelerated progress in Real-ISR, they remain suboptimal in fully leveraging model capabilities under constrained computational budgets. 


Recently, the remarkable success of Sparse Mixture-of-Experts (MoE)~\cite{shazeer2017outrageously} architectures in transformer-based large language models (LLMs), exemplified by DeepSeek~\cite{liu2024deepseek} and Grok, has reinvigorated advancements in deep learning. The MoE framework is grounded in a conceptually elegant principle: decomposing models into specialized subnetworks (``experts") optimized for distinct data patterns or tasks. By dynamically activating only task-relevant experts per input, this paradigm retains access to extensive domain expertise while preserving computational efficiency. Inspired by this success, we explore the application of sparse MoE architectures to Real-ISR tasks, where input samples exhibit substantial heterogeneity in both degradation types (e.g., blur and noise) and severity levels.

In this paper, we integrate MoE with LoRA to propose a novel mixture-of-ranks (MoR) architecture for one-step Real-ISR tasks. Instead of simply dividing each LoRA module into experts, we treat each rank within the LoRA decomposition as an independent expert, allowing the model to capture fine-grained differences and relationships within the data more effectively. This design also facilitates more flexible decomposition and recombination of expert knowledge, thereby enhancing the model’s learning capacity. Additionally, we isolate fixed-position ranks as shared experts to capture common features and reduce routing complexity. To ensure degradation-aware expert activation, we design a degradation estimation module that computes relative degradation scores using CLIP's cross-modal alignment capabilities. This module calculates cosine similarity between low-resolution (LR) images and predefined multi-dimensional positive/negative text prompts, generating adaptive routing weights. Additionally, we introduce zero-expert slots and a degradation-aware load-balancing loss, which dynamically scales the number of active experts based on input degradation severity: severely degraded samples engage more experts for enhanced reconstruction, while simpler cases utilize fewer. This adaptive scaling ensures robust restoration quality across diverse input conditions. As demonstrated in Figure~\ref{fig:teaser}, our framework achieves single-step generation of high-resolution images with enhanced fidelity. Furthermore, when compared to the multi-step Real-ISR method SeeSR,  MoR-DASR achieves a 40$\times$ speedup in inference time while maintaining comparable reconstruction quality. Our main contribution can be summarized as follows:

\begin{itemize}[leftmargin=*]

\item {We explore the integration of sparse MoE architectures into Real-ISR tasks, introducing a novel MoE architecture for single-step Real-ISR tasks that achieves high-fidelity reconstruction while maintaining resource efficiency.}

\item {We propose a Mixture-of-Ranks (MoR) architecture that designates each LoRA rank as an independent expert, enabling dynamic knowledge recombination, while isolating fixed-position ranks as shared experts to capture common knowledge and mitigate redundancy in routed experts.}

\item {We design a degradation estimation module to dynamically guide expert activation. This module leverages the cross-modal alignment capabilities to derive relative degradation scores of inputs. These scores are then integrated into the expert routing module to activate the relevant experts.}

\item {Furthermore, we introduce zero experts and design a degradation-aware load-balancing loss, which ensures dynamic allocation of computational resources based on the severity of degradation in each sample.}

\end{itemize}

\section{Related Works}

\begin{figure*}
   \centering
   \includegraphics[width=0.95\linewidth]{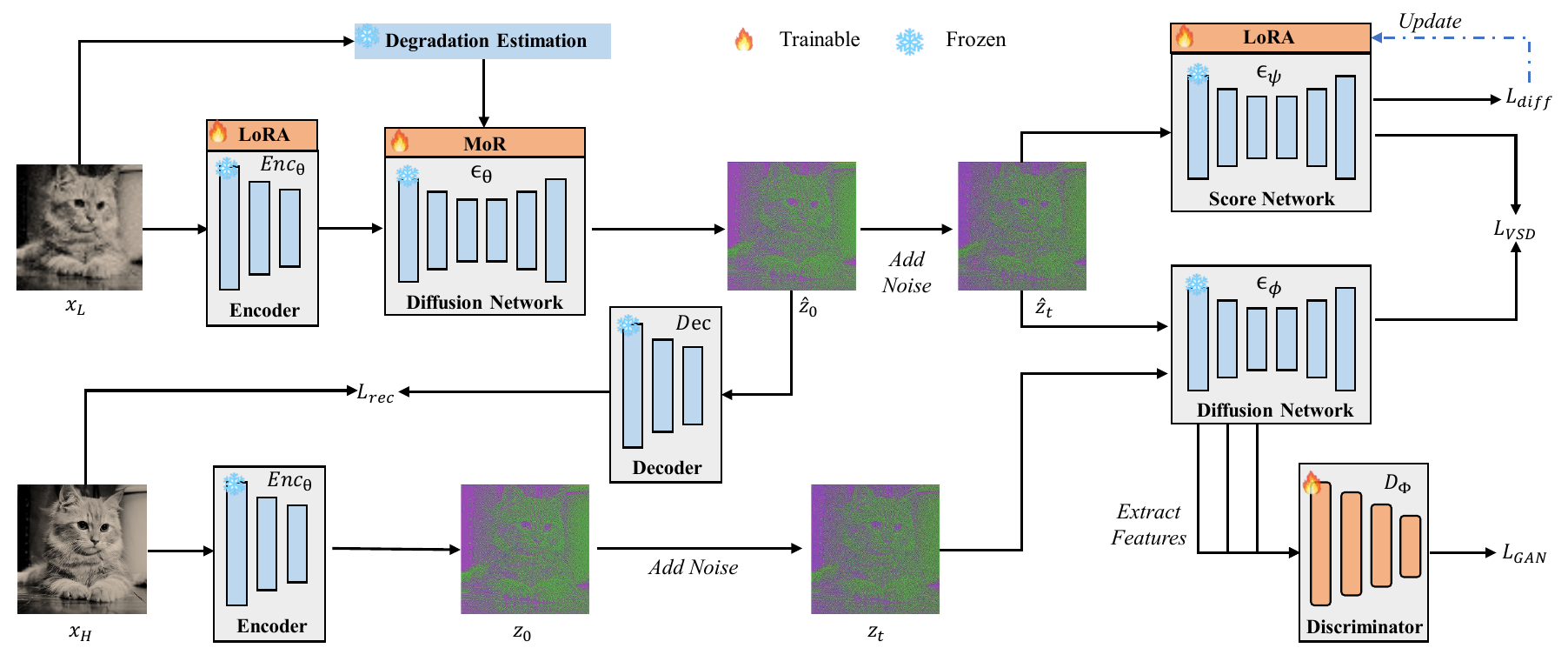}
   \caption{The training framework of MoR-DASR. The LR image is passed through a trainable encoder $Enc_{\theta}$, a diffusion network with a MoR module $\epsilon_{\theta}$ and a frozen decoder $Dec$ to obtain the desired HR image. The training procedure alternates between two phases: 1. Optimizing the variational score network $\epsilon_{\psi}$ through diffusion loss $\mathcal{L}_{diff}$ to fit the distribution of the generated samples. 2. Finetune the diffusion model $\epsilon_{\theta}$ and encoder $Enc_{\theta}$ to generate high-quality samples through reconstruction loss $\mathcal{L}_{rec}$, variational score distillation loss $\mathcal{L}_{VSD}$, and GAN loss $\mathcal{L}_{GAN}$. }
   \label{fig:framework}
\end{figure*}

\subsection{Mixture of Experts}
The Mixture of Experts (MoE) paradigm, initially introduced in~\cite{jacobs1991adaptive,jordan1994hierarchical}, has undergone significant refinement through subsequent research~\cite{collobert2001parallel,aljundi2017expert}. In MoE architecture, routers dynamically select specialized parameter subsets to process input tokens, with outputs aggregated to form the final output. This sparse activation paradigm remains prevalent in contemporary architectures and has proven instrumental in scaling large language models (LLMs). DeepSeek-MoE~\cite{liu2024deepseek} further advanced this sparse activation paradigm through two innovations: (1) Fine Expert Partitioning: Dividing the model into a larger number of specialized sub-networks. (2) Shared Expert Isolation: Designating a subset of experts to capture domain-invariant knowledge, reducing routing redundancy. These refinements have injected vitality into the development of LLMs. Even some works have extended sparse MoE to visual tasks—for instance, DiT-MoE~\cite{fei2024scaling} scales diffusion models to 16B parameters via 2 shared and 8 routed experts, advancing image generation. Concurrently, multi-task learning studies~\cite{dou2023loramoe,liu2023moelora,zhao2025each} integrate LoRA with MoE architectures, enhancing downstream task performance without proportional computational overhead.

\subsection{Real-world Image Super-Resolution}
Early Real-ISR approaches~\cite{ledig2017photo,tu2023toward,tu2024ipt} leveraged generative adversarial networks (GANs), combining adversarial and perceptual losses~\cite{zhang2018unreasonable,ding2020image} to ensure reconstruction fidelity and quality. However, the inherent complexity of real-world degradation processes causes existing methods to struggle with satisfactory restoration of severely degraded samples. Recently, diffusion models~\cite{ho2020denoising,song2020denoising,ding2025rass} has gradually emerged as a successor to GANs~\citep{goodfellow2020generative} in various downstream tasks. StableSR~\cite{wang2024exploiting} introduces an auxiliary encoder to project low-resolution (LR) features into the latent space of a pre-trained text-to-image diffusion model, fine-tuning the architecture for Real-ISR tasks. Subsequent studies~\cite{yang2023pixel,wu2024seesr,yue2024resshift} systematically investigated methods for injecting LR image information into diffusion models. Despite the improved perceptual quality, these approaches suffer from computational inefficiency, often requiring 50–100 iterative sampling steps during inference. To address this, recent work~\cite{he2024one,wang2023sinsr,wu2024one,xie2024addsr,zhang2024degradation,wang2024hero} achieves single-step super-resolution via knowledge Distillation or fine-tuning pre-trained diffusion models. However, the significant input heterogeneity inherent to Real-ISR—spanning diverse degradation types and severity levels—limits the effectiveness of these computationally intensive methods, as they fail to fully utilize model capacity under constrained computational budgets.

\section{Methodology}

\subsection{Preliminaries}

\textbf{Problem Modeling.} Real-ISR task aims to reconstruct HR images $\hat{x}_{H}$ form LR input $x_{L}$. Given a pre-trained text-to-image diffusion model, existing methods fine-tune the model to adapt to Real-ISR tasks under paired data supervision. The diffusion model receives a noisy version $z_t$ of the latent representation $z_0$ encoded by the HR image, with the condition of LR images or features extracted from it. The model is then optimized to accurately predict the noise in the latent code at each time step, which can be represented as:

\begin{equation}
    \mathcal{L} = \mathbb{E}_{z_0,t,x_L,\epsilon}\left[ \epsilon - \epsilon_{\theta}\left(z_t,t,x_L\right)\right],
    \label{eq:diffusion_loss}
\end{equation}
where $z_t = \sqrt{\alpha_t}z_0 + \sqrt{1-\alpha_t}\epsilon$, $\epsilon \in \mathcal{N} (0,I)$. During inference, the model takes gaussian noise $z_T$ as input and iteratively transforms it to the clean latent codes $\hat{z}_0$. To accelerate the sampling process, existing ISR methods employ distillation to generate the clean latent code $\hat{z}_0$ in a single step. The computation of $\hat{z}_0$ is formulated as:

\begin{equation}
    \hat{z}_0 = \frac{z_T-\sqrt{1-\alpha_T}\epsilon_{\theta}(z_T,x_L,T)}{\sqrt{\alpha_T}}.
    \label{eq:predict_x0}
\end{equation}

While these methods accelerate diffusion model inference, the random noise inherent to its input may degrade fidelity in Real-ISR tasks. To mitigate this, OSEDiff directly fine-tunes the pre-trained diffusion model to learn the LR to HR mapping, formulated as:

\begin{equation}
    \hat{z}_0 = \frac{z_L-\sqrt{1-\alpha_T}\epsilon_{\theta}(z_L,c,T)}{\sqrt{\alpha_T}},
    \label{eq:osediff}
\end{equation}
where $c$ is the prompt obtained by applying text prompt extractor DAPE to LR input. We follow this paradigm to finetune the pre-trained diffusion model to adapt Real-ISR tasks.

\textbf{Variational Score Distillation.} Variational Score Distillation (VSD) leverages the priors of a pretrained text-to-image diffusion model to optimize generative models, ensuring that generated images align semantically with the input text prompts. 

Within the VSD framework, the generator's output $\hat{z}_0$ is re-noised and fed into both a pre-trained diffusion model $\epsilon_{\phi}$ and an online-trained variational score network $\epsilon_{\psi}$. The generator is optimized by minimizing the discrepancy between the two models' predictions. As formalized in \cite{wang2023prolificdreamer}, this process is expressed as:
\begin{equation}
    \nabla_\theta \mathcal{L}_{VSD} = \mathbb{E}_{t,\epsilon} \left[\omega(t)(\epsilon_{\phi}(\hat{z}_t,t,c)-\epsilon_{\psi}(\hat{z}_t,t,c))\frac{\partial{{\hat{z}}_t}}{\partial{\theta}}\right],
    \label{eq:vsd}
\end{equation}
where $\hat{z}_t = \sqrt{\alpha_t}\hat{z}_0 + \sqrt{1-\alpha_t}\epsilon$, $\omega_t$ is a weighting function.

\subsection{Overview of MoR-DASR}

The framework of MoR-DASR is illustrated in Figure~\ref{fig:framework}. Given an LR input, the degradation estimation module first computes its degradation score. This score is fed into the dynamic routing mechanism of the Mixture-of-Ranks (MoR) module, which activates corresponding experts (ranks) to reconstruct the HR image. During fine-tuning of the diffusion model, we employ reconstruction losses (e.g., L2 and LPIPS) to enforce visual similarity between the output and the ground truth. Additionally, we align the output's data distribution with high-quality image distributions using priors from the pre-trained diffusion model, implemented via Variational Score Distillation and GAN losses.



\begin{figure}
   \centering
   \includegraphics[width=0.95\linewidth]{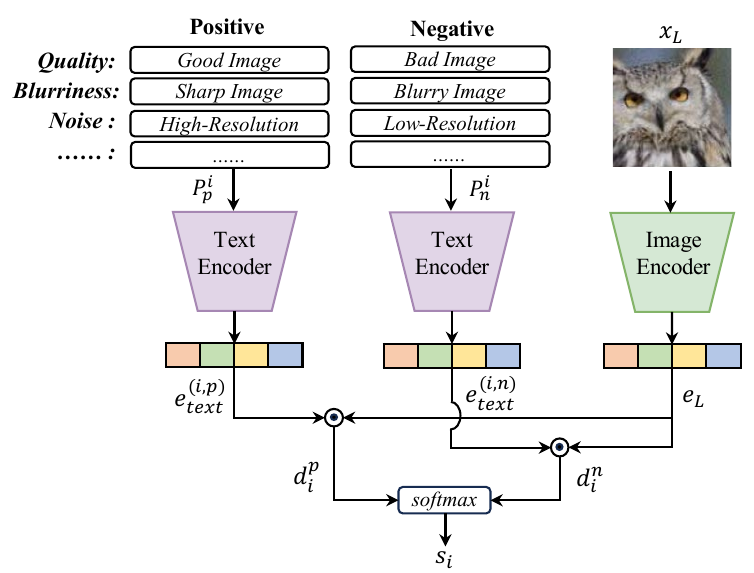}
   \caption{The framework of degradation estimation module.}
   \label{fig:deg_framework}
\end{figure}

\subsection{Degradation Estimation Module}

We integrates dynamic computation—inherent to MoE—into super-resolution tasks to enhance model performance while maintaining computational efficiency. A critical challenge lies in determining how to categorize input data for the RealISR task, specifically identifying optimal criteria to guide the expert routing mechanism. A naive approach involves routed experts based on LR input features alone. However, semantic-driven data partitioning proves suboptimal for Real-ISR, as empirical studies~\cite{liang2022efficient,zhang2024degradation} emphasize the critical role of degradation characteristics. Through empirical analysis, we observe that reconstruction quality is inversely correlated with degradation severity: images with mild degradation achieve superior reconstruction, whereas severe degradation images present greater reconstruction challenges. This finding motivates our proposal of degradation intensity as a critical criterion for dynamic computation in expert-based super-resolution frameworks.

Based on the above analysis, we utilize the cross-modality ability of CLIP model and proposed a novel degradation estimation module. We first defined a set of positive and negative prompt pairs from multiple perspectives for evaluating image quality, which includes the overall quality of the image, as well as the degree of blur, whether it contains noise, and so on. Then, for each input image, we obtain its embedding $e_{L}$ through CLIP image encoder:

\begin{equation}
    e_L = f_{img}(x_L).
    \label{eq:clip_img}
\end{equation}

Then, we calculate the embedding vectors corresponding to the positive and negative prompts on each evaluation dimension:

\begin{equation}
    e_{text}^{(i,p)} = f_{text}(P_p^i),  ~~~~~e_{text}^{(i,n)} = f_{text}(P_n^i),
    \label{eq:clip_text}
\end{equation}
where $ e_{text}^{(i,p)}$ and $ e_{text}^{(i,n)}$ denote the embeddings of the positive $(P_p^i)$ and negative prompts $(P_n^i)$ of the i-th evaluation dimension, respectively. We then combine image embedding with the distance between positive and negative text embeddings to obtain the degradation score:

\begin{equation}
    d^{(i,p)} = \frac{e_L\odot e^{(i,p)}_{text}}{||e_L|| \cdot ||e^{(i,p)}_{text}||}, ~~~~~ d^{(i,n)} = \frac{e_L\odot e^{(i,n)}_{text}}{||e_L|| \cdot ||e^{(i,n)}_{text}||},
    \label{eq:cosine_distance}
\end{equation}

\begin{equation}
    s_{i} = \frac{\exp(d^{(i,n)})}{\exp(d^{(i,p)}) + \exp(d^{(i,n)})},
    \label{eq:deg_score}
\end{equation}
where $d^{(i,p)}$ and $d^{(i,n)}$ denote the cosine distance between the LR image and positive and negative prompts on the i-th evaluation dimension,respectively. $s_i$ represents the final degradation score on the i-th evaluation dimension.

\begin{figure}
    \centering
    \includegraphics[width=0.95\linewidth]{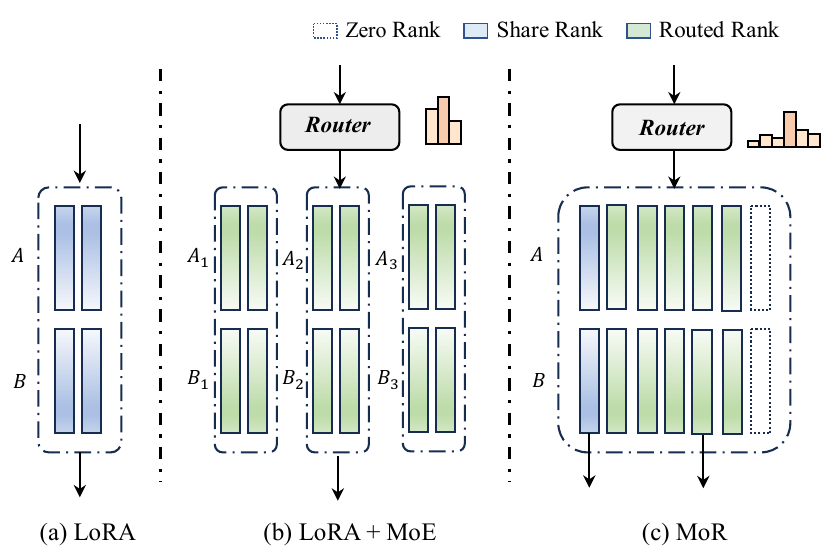}
    \caption{Comparison of LoRA, LoRA MoE and MoR. In MoR, each rank is treated as an expert. A subset of these ranks is designated as shared experts to process all samples, while the remaining ranks function as routed experts that are selectively activated to process specific samples.}
    \label{fig:moe}
    \vspace{-2mm}
 \end{figure}

\subsection{MoR with Degradation-Aware Router}
Existing Parameter-Efficient Fine-Tuning (PEFT) methods treat each LoRA module as a standalone expert, assigning individual experts to specific tasks (as shown in Figure ~\ref{fig:moe}). However, this static partitioning of the parameter space restricts the dynamic fusion and decomposition of expert knowledge, as entire modules are activated or deactivated holistically—a coarse-grained strategy that results in suboptimal utilization of learned features. In contrast, we treat each rank of the LoRA as an independent expert, enabling fine-grained expert activation and allowing knowledge to be flexibly decomposed and recombined.

The architecture of MoR is illustrated in Figure~\ref{fig:moe}. Building upon a pre-trained text-to-image diffusion model, we integrate trainable low-rank matrices $A \in R^{ d\times r} $ and $B \in R^{ r \times d} $ to fine-tune the model parameters, where $d$ denotes the pre-trained weights' channel dimension, $r$ denotes the rank of the LoRA module. We treat each rank as an independent expert, partitioned into two categories: shared experts and routed experts. Routed experts are activated via gating mechanisms and top-k selection strategies, whereas shared experts remain persistently active to capture common knowledge. Specifically, the degradation score $s$ of input LR images, computed by the degradation estimation module, is fed into a gating function to generate logits. Experts corresponding to the top-k logits are subsequently activated. This process is formalized as follows:

\begin{equation}
    g_i(s) = TopK(Softmax(sW_g)),
    \label{eq:gating}
\end{equation}
where $W_g$ is a learnable gating matrix,  $g_i(s)$ denotes the logits of the selected experts $i$. Following the selection of $k$ experts, the forward pass of MoR is formulated as: 

\begin{equation}
    MoR(z_t) = W_0 x + \sum_{i=1}^k g_i(s)B_i A_ix + \sum_{j=1}^m B_j A_j x,
    \label{eq:gating2}
\end{equation}
where $W_0$ represents the parameter matrix of the backbone model, $m$ is the number of shared experts.

Building on the aforementioned framework, we introduce zero experts to enable dynamic rank adaptation, addressing two key challenges: (1) input samples with varying degradation severity necessitate distinct computational budgets, and (2) the optimal LoRA rank may differ across model layers. For mildly degraded samples, the network activates zero experts—effectively bypassing unnecessary computations—to prevent over-restoration. Conversely, severely degraded inputs trigger the activation of more real experts, allocating greater computational resources to achieve high-fidelity reconstruction. This adaptive mechanism ensures efficient resource distribution, balancing restoration quality and computational cost across diverse degradation scenarios.

\begin{figure*}
    \centering
    \includegraphics[width=0.95\linewidth]{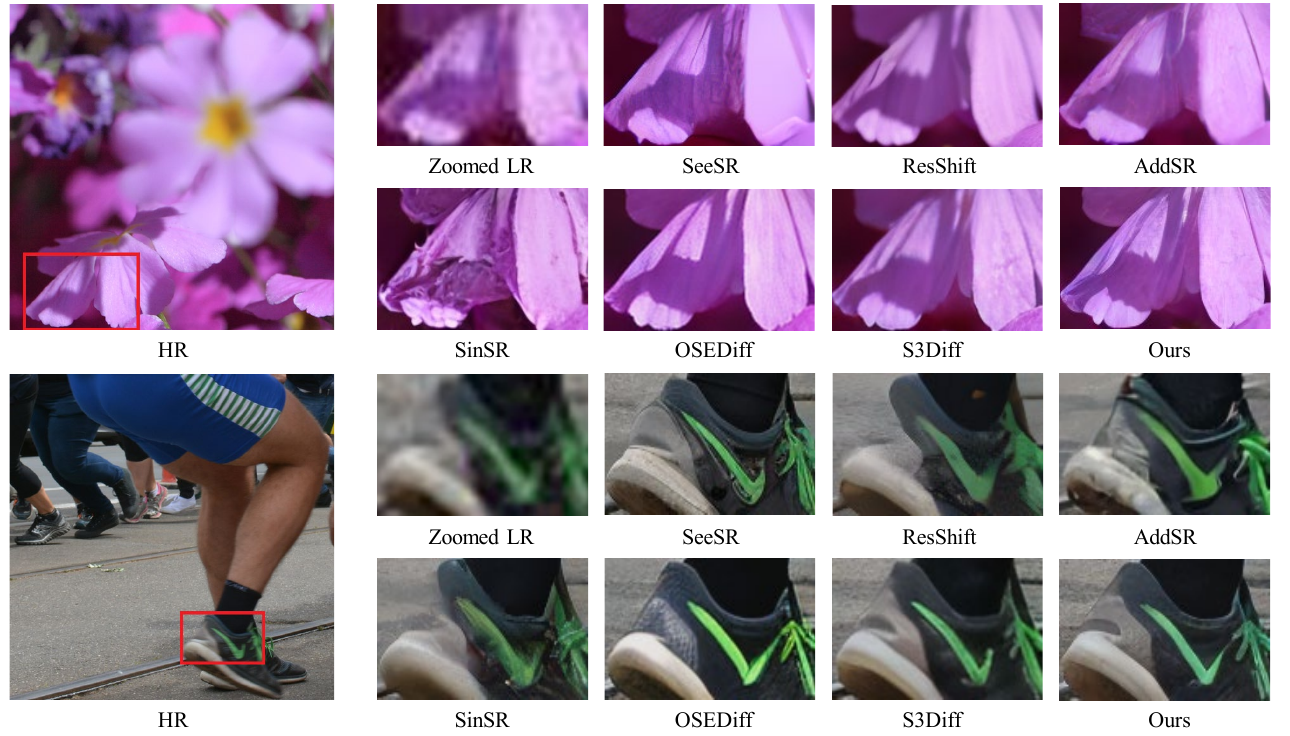}
    \caption{Visual comparisons of different Real-ISR methods. Please zoom in for a better view.}
    \label{fig:results}
 \end{figure*}

\subsection{Loss Functions}
To enable single-step high-quality image reconstruction, we employ a composite loss function comprising reconstruction loss and feature-distribution matching loss. The reconstruction loss combines L2 loss and LPIPS loss to enforce structural and perceptual fidelity. The feature-distribution matching loss integrates two components: (1) VSD loss, which aligns the model's output distribution with the high-quality prior embedded in the pre-trained text-to-image diffusion model, and (2) GAN loss, leveraging the diffusion model's feature extraction capability to align outputs with real data distributions under varying noise perturbations through adversarial training. Furthermore, we introduce a degradation-aware load-balancing loss to mitigate suboptimal resource allocation and improve overall computational efficiency. Thus, the full optimization objective can be expressed as follows:

\begin{equation}
    \mathcal{L}_\theta = \mathcal{L}_{rec} + \lambda_{1}\mathcal{L}_{VSD} + \lambda_2 \mathcal{L}_{GAN} + \mathcal{L}_{balance}.
    \label{eq:total loss}
\end{equation}


\textbf{GAN loss.} Following \cite{he2024one}, we leverage a pre-trained diffusion model to extract features from both synthetic and real data under varying noise perturbations. These features are then processed by multi-scale tiny discriminator heads $D_{\Phi}$ to distinguish between the two domains. The adversarial training objective is formalized as:

\begin{equation}
\mathcal{L}_{adv}^{\epsilon_{\theta}} = -\mathbb{E}_{\hat{z}_0} \left[ D_{\Phi}\left(\hat{z}_{t},t\right) \right],
\label{eq:adv_G}
\end{equation}
where $\hat{z}_{t}$ denotes the latent representation derived by re-adding noise to the generator's initial output.

\textbf{Degradation-aware load-balancing loss.}
In MoE's training process, load-balancing loss plays a critical role. It penalizes imbalanced expert selection and encourages equitable utilization of experts. It prevents the model from over-relying on specific experts, thereby mitigating suboptimal resource allocation and improving overall computational efficiency, which can be formalized as:

\begin{equation}
\mathcal{L}_{balance} = \alpha N \sum_{i=1}^{N} f_i ~ \mathcal{P}_{i},
\label{eq:balance loss}
\end{equation}
where $N$ is the number of routed experts, $\alpha$ is a weighting factor and $f_i$ is the fraction of samples dispatched to expert $i$:

\begin{equation}
f_{i} = \frac{1}{b} \sum_{x \in \mathcal{B}} 1 
\left\{\text{argmax}~p(x) = i\right\},
\label{eq:fi}
\end{equation}
where $\mathcal{B}$ is the sample batch, $b$ is the batch size, $p(x)$ denotes the probability value of the input after gating calculation and $\mathcal{P}_{i}$ is the fraction of router probability allocated for expert $i$:

\begin{equation}
\mathcal{P}_{i} = \frac{1}{b} \sum_{x \in \mathcal{B}}p_i(x).
\label{eq:pi}
\end{equation}

However, the formulation treats zero and real experts homogeneously for load balancing, assigning zero experts a static activation probability of approximately $k/N$, independent of input degradation severity. This violates our core objective: adaptive expert allocation, where simple samples activate more zero experts (minimizing computation) and complex samples prioritize real experts (maximizing reconstruction quality). To resolve this, we propose a degradation-aware load-balancing loss by revising Eq.~\ref{eq:balance loss} to incorporate degradation severity into the balancing criterion, ensuring expert activation aligns dynamically with input complexity.

\begin{equation}
\mathcal{L}_{balance} = N \sum_{i=1}^{N} \alpha_i ~ f_i ~ p_i,
\label{eq:balance loss2}
\end{equation}
where
\begin{align}
    \alpha_i = 
    \begin{cases}
        \alpha & \text{if } i \leq n, \\
        s \alpha & \text{if } i> n,
    \end{cases}
    \label{eq:deg-banlance_loss}
\end{align}
where $n$ is the number of true experts and $s$ is the degradation score of the input sample. It can be seen that the degradation score $s$ (higher values indicate higher degradation) modulates the penalization weight for zero experts. Specifically, higher degradation (larger $s$) increases the penalty for zero expert activation, thereby incentivizing the model to prioritize real experts. Conversely, for mildly degraded inputs (lower $s$), the reduced penalty encourages zero expert usage, efficiently allocating computational resources without compromising restoration quality.

\begin{table*}[t]
    \centering
    \begin{tabular}{c|c|ccccccccc}
        \toprule
         Datasets & Methods  &PSNR$\uparrow$ &SSIM$\uparrow$ &LPIPS $\downarrow$   &CLIPIQA$\uparrow$ & MUSIQ$\uparrow$   & MANIQA$\uparrow$ & TOPIQ$\uparrow$  & TRES$\uparrow$ \\
            \midrule
           \multirow{5}*{DIV2K-Val} 
            &AddSR    & 23.26 &0.590 & 0.362 & 0.573 &63.39 &0.405 &0.573 &73.23  \\
            &SinSR & \textbf{24.41} & 0.602 & 0.324  & 0.648 & 62.80 & 0.424  &0.571 &72.24 \\
            &OSEDiff &23.92 &\textbf{0.614} &0.310 & 0.659 & 67.71  & 0.435  &0.606   &78.40  \\
            &S3Diff   & 23.53 &0.593 & \textbf{0.258} & \textbf{0.699} & \underline{67.92} &\underline{0.452}  & \underline{0.633} & \underline{80.72} \\
            &MoR-DASR &\underline{24.01} &\underline{0.606} &\underline{0.289} &\underline{0.681} &\textbf{68.09} &\textbf{0.475} &\textbf{0.663} &\textbf{84.14} \\ 
            \midrule
            \multirow{5}*{RealSR}
            &AddSR    & 23.12 &0.655 & 0.309 & 0.552 &67.14 &\underline{0.488} &0.599 &79.91  \\
            &SinSR & \textbf{26.16} & \textbf{0.739} & 0.308  & 0.631 & 60.96 & 0.399  &0.512 &59.92 \\
            &OSEDiff & 25.26 &\underline{0.728} &0.301 & 0.651 & \underline{68.41}  & 0.468 &0.614 &\underline{80.18}\\
            &S3Diff   & 25.18 &0.727 & \textbf{0.272}  & \underline{0.672} & 67.82 &0.459 &\underline{0.616} &78.82 \\
            &MoR-DASR  &\underline{25.32} & \underline{0.728} & \underline{0.291} &\textbf{0.691} &\textbf{69.78} & \textbf{0.512}  & \textbf{0.662} & \textbf{84.97} \\   
            \midrule
            \multirow{5}*{DRealSR}
            &AddSR    & 26.71 &0.738 & 0.321 & 0.593 &62.13 &0.458  &0.569 &71.41  \\
            &SinSR & \underline{28.32} & 0.747 & 0.372  & 0.642 & 55.36 & 0.388 &0.512 &59.92 \\
            &OSEDiff &28.29 &\textbf{0.792} & \textbf{0.302}  & 0.673 & \underline{64.47}  & \underline{0.469} &\underline{0.616} & \underline{76.76} \\
            &S3Diff   & 27.54 &0.767 & 0.311 & \underline{0.702} & 63.94 &0.452  &0.604 &75.41\\
            &MoR-DASR  &\textbf{28.37} &\underline{0.776} &\underline{0.307} &\textbf{0.717} & \textbf{65.94} & \textbf{0.509}   & \textbf{0.652} & \textbf{81.78}\\ 
      \bottomrule
    \end{tabular} 
    \caption{Quantitative comparison with state of the arts one-step Real-ISR methods. The best and second best results are highlighted in \textbf{bold} and \underline{underline}.}
    \label{tab:comparison_one_step}
\end{table*}

\section{Experiments}
\subsection{Experiments Setup}

\textbf{Training Details.} Following SeeSR~\cite{wu2024seesr}, we utilize LSDIR~\cite{li2023lsdir} dataset and 10k face images from FFHQ~\cite{karras2019style} dataset for training. To generate low-resolution (LR) and high-resolution (HR) training pairs, we apply the degradation pipeline proposed in Real-ESRGAN. We employ Stable Diffusion 2.1 as the base model and fine-tune it to adapt Real-ISR tasks. The rank parameter for LoRA is set to 4 in both the VAE encoder and the variational score network. The Mixture-of-Ranks (MoR) module comprises 40 ranks, including 8 shared and 32 routed ranks, with a top-8 routing strategy implemented during training. The model undergoes 25,000 iterations with a batch size of 16 and a learning rate of $5e-5$.

\textbf{Compared Methods.}
We compare MoR-DASR with state-of-the-art one-step Real-ISR methods, including SinSR~\cite{wang2023sinsr}, AddSR~\cite{xie2024addsr}, OSEDiff~\cite{wu2024one}, and S3Diff~\cite{zhang2024degradation}. Additionally, to comprehensively evaluate the performance of MoR-DASR, we also compare it with multi-step Real-ISR methods~\cite{lin2023diffbir,wang2024exploiting,yue2024resshift,wu2024seesr} and GAN-based Real-ISR approaches~\cite{zhang2021designing,wang2021real,liang2022details,chen2022femasr}. Experimental details are provided in the appendix.

\textbf{Testing Details.}
We evaluate the performance of Real-ISR algorithms on two real datasets, RealSR~\cite{cai2019toward} and DRealSR~\cite{wu2024seesr}, and one synthetic dataset, DIV2K-Val~\cite{agustsson2017ntire}. We adopt non-reference metrics, including CLIPIQA~\cite{wang2022exploring}, MUSIQ~\cite{ke2021musiq}, MANIQA~\cite{yang2022maniqa}, TOPIQ~\cite{chen2024topiq}, and TRES~\cite{golestaneh2022no} to evaluate the performance of various Real-ISR methods. In Real-ISR tasks, these metrics are crucial as they better align with human visual perception~\cite{wang2024exploiting,xie2024addsr}. Additionally, we also report traditional metrics such as PSNR, SSIM~\cite{wang2004image}, and LPIPS~\cite{zhang2018unreasonable} for reference.

\subsection{Comparison with State-of-the-Arts}

\textbf{Quantitative Comparison}
Tables~\ref{tab:comparison_one_step} present a quantitative comparison of state-of-the-art methods on three benchmark datasets. Table~\ref{tab:comparison_one_step} primarily evaluates MoR-DASR against existing one-step Real-ISR approaches. The following key observations can be drawn from the table: (1) MoR-DASR consistently ranks among the top-performing methods across all metrics and datasets. Notably, on the DrealSR dataset, it achieves the highest scores for nearly every metric, underscoring its superior performance. (2) While SinSR distills a multi-step Real-ISR model trained from scratch and delivers competitive results in PSNR and SSIM, its non-reference metrics are markedly inferior to other single-step methods. (3) S3diff attains the best LPIPS scores, particularly on the synthetic DIV2K dataset, but its performance on other metrics was suboptimal. This may be attributed to its substantial emphasis on minimizing LPIPS loss during training. (4) Compared to other single-step methods, our approach demonstrates significant improvements in non-reference metrics (e.g., MANIOA, TOPIQ). These non-reference metrics prioritize perceptual quality and are closely aligned with human perceptual evaluations, further validating MoR-DASR's ability to generate reconstructions that align with human visual preferences.

\textbf{Qualitative Comparison}
Figure \ref{fig:results} present qualitative comparisons of reconstruction results. Our analysis demonstrates that existing methods frequently struggle to recover fine-grained details, whereas our method generates clear and accurate textures. While some baselines produce sharp details, these often deviate from ground-truth structures—for instance, synthesizing erroneous shoe logos. In contrast, our method not only reconstructs complex details but also avoids generating unnatural artifacts, achieving a balance between high-quality texture generation and visual fidelity. Extended visualizations are provided in the appendix.

\begin{table}[t]
    \centering
    \begin{tabular}{c|cccc}
    \Xhline{0.8pt}
    Model   & CLIPIQA $\uparrow$  & MANIOA $\uparrow$ & TRES $\uparrow$\\
    \Xhline{0.4pt}
    LoRA  & 0.670 & 0.481 & 78.81 \\
    LoRA+MoE & 0.689 & 0.484  & 79.36 \\
    MoR-v1   & 0.704 & 0.491  & 80.32  \\
    MoR-v2   & 0.699 & 0.479 & 79.41 \\
    MoR-full  & \textbf{0.717} & \textbf{0.509} &\textbf{81.78} \\
    \Xhline{0.4pt}
    \end{tabular} 
    \caption{Ablation study of our proposed MoR architecture.}
    \label{tab:ablation_mor}
 \end{table}

\subsection{Ablation Studies}
In this section, we evaluate the contributions of different components in our method. Specifically, we focus on analyzing the impact of the MoR architecture and the degradation-aware load balancing loss on model performance. Additional ablation studies are provided in the appendix.

Table~\ref{tab:ablation_mor} presents an ablation study evaluating the efficacy of our MoR module. We compare five configurations on the real-world SR dataset. (1) Vanilla LoRA. (2) LoRA+MoE: Integration of LoRA with a standard MoE architecture. (3) MoR-v1: MoR without zero experts. (4) MoR-v2: MoR with zero experts but excluding degradation-aware load balancing loss. (5) MoR-full: our proposed MoR framework. According to the Table, we have key findings as follows. The MoE architecture (LoRA+MoE) improves baseline performance, validating the effectiveness of applying MoE in Real-ISR tasks. By designing each rank in LoRA as an independent expert (MoR-v1), the model is able to flexibly decompose and recombine knowledge, further enhancing performance. MoR-v2 introduces zero experts but still relies on a traditional load balancing loss to guide expert selection, resulting in a significant drop in performance. In contrast, the MoR-full model incorporates a degradation-aware load balancing loss while introducing zero experts, achieving SOTA results: $+7\%$ CLIPIQA,  $+5.8\%$ MANIQA, and  $+3.8\%$ TRES over baselines. It demonstrates that our method significantly enhances the quality of reconstructed images.



\section{Conclusion}
We present MoR-DASR, a novel MoE architecture for one-step Real-ISR tasks. MoR-DASR captures common knowledge through shared rank in a mixture-of-ranks architecture, decomposing and recombining features via routed ranks to enhance model performance without sacrificing computational efficiency. To guide the MoR module in selecting input-relevant experts, we design a degradation estimation module that leverages CLIP's cross-modal alignment capability, mapping sample-specific degradation scores to the router's gating mechanism. Additionally, we introduce a degradation-aware load-balancing loss combining zero experts, which dynamically adjusts the number of activated ranks (experts) to optimize reconstruction quality across inputs with diverse degradation levels. Extensive experiments demonstrate that MoR-DASR achieves state-of-the-art performance while retaining practical inference efficiency.

{
    \small
    \bibliographystyle{ieeenat_fullname}
    \bibliography{main}

@article{ding2020image,
  title={Image quality assessment: Unifying structure and texture similarity},
  author={Ding, Keyan and Ma, Kede and Wang, Shiqi and Simoncelli, Eero P},
  journal={IEEE transactions on pattern analysis and machine intelligence},
  volume={44},
  number={5},
  pages={2567--2581},
  year={2020},
  publisher={IEEE}
}

@inproceedings{wang2021real,
  title={Real-esrgan: Training real-world blind super-resolution with pure synthetic data},
  author={Wang, Xintao and Xie, Liangbin and Dong, Chao and Shan, Ying},
  booktitle={Proceedings of the IEEE/CVF international conference on computer vision},
  pages={1905--1914},
  year={2021}
}

@article{wang2024exploiting,
  title={Exploiting diffusion prior for real-world image super-resolution},
  author={Wang, Jianyi and Yue, Zongsheng and Zhou, Shangchen and Chan, Kelvin CK and Loy, Chen Change},
  journal={International Journal of Computer Vision},
  pages={1--21},
  year={2024},
  publisher={Springer}
}

@article{yang2023pixel,
  title={Pixel-aware stable diffusion for realistic image super-resolution and personalized stylization},
  author={Yang, Tao and Ren, Peiran and Xie, Xuansong and Zhang, Lei},
  journal={arXiv preprint arXiv:2308.14469},
  year={2023}
}

@article{yue2024resshift,
  title={Resshift: Efficient diffusion model for image super-resolution by residual shifting},
  author={Yue, Zongsheng and Wang, Jianyi and Loy, Chen Change},
  journal={Advances in Neural Information Processing Systems},
  volume={36},
  year={2024}
}

@article{xie2024addsr,
  title={AddSR: Accelerating Diffusion-based Blind Super-Resolution with Adversarial Diffusion Distillation},
  author={Xie, Rui and Tai, Ying and Zhang, Kai and Zhang, Zhenyu and Zhou, Jun and Yang, Jian},
  journal={arXiv preprint arXiv:2404.01717},
  year={2024}
}

@article{wang2023sinsr,
  title={SinSR: Diffusion-Based Image Super-Resolution in a Single Step},
  author={Wang, Yufei and Yang, Wenhan and Chen, Xinyuan and Wang, Yaohui and Guo, Lanqing and Chau, Lap-Pui and Liu, Ziwei and Qiao, Yu and Kot, Alex C and Wen, Bihan},
  journal={arXiv preprint arXiv:2311.14760},
  year={2023}
  
}

@inproceedings{wang2018esrgan,
  title={Esrgan: Enhanced super-resolution generative adversarial networks},
  author={Wang, Xintao and Yu, Ke and Wu, Shixiang and Gu, Jinjin and Liu, Yihao and Dong, Chao and Qiao, Yu and Change Loy, Chen},
  booktitle={Proceedings of the European conference on computer vision (ECCV) workshops},
  pages={0--0},
  year={2018}
}

@inproceedings{zhang2021designing,
  title={Designing a practical degradation model for deep blind image super-resolution},
  author={Zhang, Kai and Liang, Jingyun and Van Gool, Luc and Timofte, Radu},
  booktitle={Proceedings of the IEEE/CVF International Conference on Computer Vision},
  pages={4791--4800},
  year={2021}
}

@inproceedings{liang2021swinir,
  title={Swinir: Image restoration using swin transformer},
  author={Liang, Jingyun and Cao, Jiezhang and Sun, Guolei and Zhang, Kai and Van Gool, Luc and Timofte, Radu},
  booktitle={Proceedings of the IEEE/CVF international conference on computer vision},
  pages={1833--1844},
  year={2021}
}

@inproceedings{liang2022efficient,
  title={Efficient and degradation-adaptive network for real-world image super-resolution},
  author={Liang, Jie and Zeng, Hui and Zhang, Lei},
  booktitle={European Conference on Computer Vision},
  pages={574--591},
  year={2022},
  organization={Springer}
}

@article{ho2020denoising,
  title={Denoising diffusion probabilistic models},
  author={Ho, Jonathan and Jain, Ajay and Abbeel, Pieter},
  journal={Advances in neural information processing systems},
  volume={33},
  pages={6840--6851},
  year={2020}
}

@article{song2020denoising,
  title={Denoising diffusion implicit models},
  author={Song, Jiaming and Meng, Chenlin and Ermon, Stefano},
  journal={arXiv preprint arXiv:2010.02502},
  year={2020}
}

@article{goodfellow2020generative,
  title={Generative adversarial networks},
  author={Goodfellow, Ian and Pouget-Abadie, Jean and Mirza, Mehdi and Xu, Bing and Warde-Farley, David and Ozair, Sherjil and Courville, Aaron and Bengio, Yoshua},
  journal={Communications of the ACM},
  volume={63},
  number={11},
  pages={139--144},
  year={2020},
  publisher={ACM New York, NY, USA}
}

@inproceedings{karras2019style,
  title={A style-based generator architecture for generative adversarial networks},
  author={Karras, Tero and Laine, Samuli and Aila, Timo},
  booktitle=cvpr,
  pages={4401--4410},
  year={2019}
}

@InProceedings{zhang2018unreasonable,
  title={The unreasonable effectiveness of deep features as a perceptual metric},
  author={Zhang, Richard and Isola, Phillip and Efros, Alexei A and Shechtman, Eli and Wang, Oliver},
  booktitle=cvpr,
  pages={586--595},
  year={2018}
}

@article{wang2004image,
  title={Image quality assessment: from error visibility to structural similarity},
  author={Wang, Zhou and Bovik, Alan C and Sheikh, Hamid R and Simoncelli, Eero P},
  journal=tip,
  volume={13},
  number={4},
  pages={600--612},
  year={2004},
  publisher={IEEE}
}

@InProceedings{wang2022exploring,
    author = {Wang, Jianyi and Chan, Kelvin CK and Loy, Chen Change},
    title = {Exploring CLIP for Assessing the Look and Feel of Images},
    booktitle = aaai,
    year = {2023}
}

@InProceedings{ke2021musiq,
  title={Musiq: Multi-scale image quality transformer},
  author={Ke, Junjie and Wang, Qifei and Wang, Yilin and Milanfar, Peyman and Yang, Feng},
  booktitle=iccv,
  pages={5148--5157},
  year={2021}
}

@InProceedings{cai2019toward,
  title={Toward real-world single image super-resolution: A new benchmark and a new model},
  author={Cai, Jianrui and Zeng, Hui and Yong, Hongwei and Cao, Zisheng and Zhang, Lei},
  booktitle=iccv,
  pages={3086--3095},
  year={2019}
}

@inproceedings{liang2022details,
  title={Details or artifacts: A locally discriminative learning approach to realistic image super-resolution},
  author={Liang, Jie and Zeng, Hui and Zhang, Lei},
  booktitle={Proceedings of the IEEE/CVF Conference on Computer Vision and Pattern Recognition},
  pages={5657--5666},
  year={2022}
}

@inproceedings{chen2022femasr,
  title={Real-world blind super-resolution via feature matching with implicit high-resolution priors},
  author={Chen, Chaofeng and Shi, Xinyu and Qin, Yipeng and Li, Xiaoming and Han, Xiaoguang and Yang, Tao and Guo, Shihui},
  booktitle={Proceedings of the 30th ACM International Conference on Multimedia},
  pages={1329--1338},
  year={2022}
}

@article{lin2023diffbir,
  title={DiffBIR: Towards Blind Image Restoration with Generative Diffusion Prior},
  author={Lin, Xinqi and He, Jingwen and Chen, Ziyan and Lyu, Zhaoyang and Fei, Ben and Dai, Bo and Ouyang, Wanli and Qiao, Yu and Dong, Chao},
  journal={arXiv preprint arXiv:2308.15070},
  year={2023}
}

@article{wu2024one,
  title={One-Step Effective Diffusion Network for Real-World Image Super-Resolution},
  author={Wu, Rongyuan and Sun, Lingchen and Ma, Zhiyuan and Zhang, Lei},
  journal={arXiv preprint arXiv:2406.08177},
  year={2024}
}

@inproceedings{agustsson2017ntire,
  title={Ntire 2017 challenge on single image super-resolution: Dataset and study},
  author={Agustsson, Eirikur and Timofte, Radu},
  booktitle={Proceedings of the IEEE conference on computer vision and pattern recognition workshops},
  pages={126--135},
  year={2017}
}

@inproceedings{li2023lsdir,
  title={Lsdir: A large scale dataset for image restoration},
  author={Li, Yawei and Zhang, Kai and Liang, Jingyun and Cao, Jiezhang and Liu, Ce and Gong, Rui and Zhang, Yulun and Tang, Hao and Liu, Yun and Demandolx, Denis and others},
  booktitle={Proceedings of the IEEE/CVF Conference on Computer Vision and Pattern Recognition},
  pages={1775--1787},
  year={2023}
}

@inproceedings{wu2024seesr,
  title={Seesr: Towards semantics-aware real-world image super-resolution},
  author={Wu, Rongyuan and Yang, Tao and Sun, Lingchen and Zhang, Zhengqiang and Li, Shuai and Zhang, Lei},
  booktitle={Proceedings of the IEEE/CVF conference on computer vision and pattern recognition},
  pages={25456--25467},
  year={2024}
}

@inproceedings{yang2022maniqa,
  title={Maniqa: Multi-dimension attention network for no-reference image quality assessment},
  author={Yang, Sidi and Wu, Tianhe and Shi, Shuwei and Lao, Shanshan and Gong, Yuan and Cao, Mingdeng and Wang, Jiahao and Yang, Yujiu},
  booktitle={Proceedings of the IEEE/CVF Conference on Computer Vision and Pattern Recognition},
  pages={1191--1200},
  year={2022}
}

@article{he2024one,
  title={One step diffusion-based super-resolution with time-aware distillation},
  author={He, Xiao and Tang, Huaao and Tu, Zhijun and Zhang, Junchao and Cheng, Kun and Chen, Hanting and Guo, Yong and Zhu, Mingrui and Wang, Nannan and Gao, Xinbo and others},
  journal={arXiv preprint arXiv:2408.07476},
  year={2024}
}

@article{zhang2015feature,
  title={A feature-enriched completely blind image quality evaluator},
  author={Zhang, Lin and Zhang, Lei and Bovik, Alan C},
  journal={IEEE Transactions on Image Processing},
  volume={24},
  number={8},
  pages={2579--2591},
  year={2015},
  publisher={IEEE}
}

@inproceedings{dong2014learning,
  title={Learning a deep convolutional network for image super-resolution},
  author={Dong, Chao and Loy, Chen Change and He, Kaiming and Tang, Xiaoou},
  booktitle={Computer Vision--ECCV 2014: 13th European Conference, Zurich, Switzerland, September 6-12, 2014, Proceedings, Part IV 13},
  pages={184--199},
  year={2014},
  organization={Springer}
}

@inproceedings{kim2016accurate,
  title={Accurate image super-resolution using very deep convolutional networks},
  author={Kim, Jiwon and Lee, Jung Kwon and Lee, Kyoung Mu},
  booktitle={Proceedings of the IEEE conference on computer vision and pattern recognition},
  pages={1646--1654},
  year={2016}
}

@article{bao2022attention,
  title={Attention-driven graph neural network for deep face super-resolution},
  author={Bao, Qiqi and Gang, Bowen and Yang, Wenming and Zhou, Jie and Liao, Qingmin},
  journal={IEEE Transactions on Image Processing},
  volume={31},
  pages={6455--6470},
  year={2022},
  publisher={IEEE}
}

@inproceedings{chen2023activating,
  title={Activating more pixels in image super-resolution transformer},
  author={Chen, Xiangyu and Wang, Xintao and Zhou, Jiantao and Qiao, Yu and Dong, Chao},
  booktitle={Proceedings of the IEEE/CVF conference on computer vision and pattern recognition},
  pages={22367--22377},
  year={2023}
}

@article{zhang2024degradation,
  title={Degradation-guided one-step image super-resolution with diffusion priors},
  author={Zhang, Aiping and Yue, Zongsheng and Pei, Renjing and Ren, Wenqi and Cao, Xiaochun},
  journal={arXiv preprint arXiv:2409.17058},
  year={2024}
}

@article{hu2022lora,
  title={Lora: Low-rank adaptation of large language models.},
  author={Hu, Edward J and Shen, Yelong and Wallis, Phillip and Allen-Zhu, Zeyuan and Li, Yuanzhi and Wang, Shean and Wang, Lu and Chen, Weizhu and others},
  journal={ICLR},
  volume={1},
  number={2},
  pages={3},
  year={2022}
}

@article{shazeer2017outrageously,
  title={Outrageously large neural networks: The sparsely-gated mixture-of-experts layer},
  author={Shazeer, Noam and Mirhoseini, Azalia and Maziarz, Krzysztof and Davis, Andy and Le, Quoc and Hinton, Geoffrey and Dean, Jeff},
  journal={arXiv preprint arXiv:1701.06538},
  year={2017}
}

@article{dai2024deepseekmoe,
  title={Deepseekmoe: Towards ultimate expert specialization in mixture-of-experts language models},
  author={Dai, Damai and Deng, Chengqi and Zhao, Chenggang and Xu, RX and Gao, Huazuo and Chen, Deli and Li, Jiashi and Zeng, Wangding and Yu, Xingkai and Wu, Yu and others},
  journal={arXiv preprint arXiv:2401.06066},
  year={2024}
}

@article{liu2024deepseek,
  title={Deepseek-v3 technical report},
  author={Liu, Aixin and Feng, Bei and Xue, Bing and Wang, Bingxuan and Wu, Bochao and Lu, Chengda and Zhao, Chenggang and Deng, Chengqi and Zhang, Chenyu and Ruan, Chong and others},
  journal={arXiv preprint arXiv:2412.19437},
  year={2024}
}

@article{jacobs1991adaptive,
  title={Adaptive mixtures of local experts},
  author={Jacobs, Robert A and Jordan, Michael I and Nowlan, Steven J and Hinton, Geoffrey E},
  journal={Neural computation},
  volume={3},
  number={1},
  pages={79--87},
  year={1991},
  publisher={MIT Press}
}

@article{jordan1994hierarchical,
  title={Hierarchical mixtures of experts and the EM algorithm},
  author={Jordan, Michael I and Jacobs, Robert A},
  journal={Neural computation},
  volume={6},
  number={2},
  pages={181--214},
  year={1994},
  publisher={MIT Press}
}

@article{collobert2001parallel,
  title={A parallel mixture of SVMs for very large scale problems},
  author={Collobert, Ronan and Bengio, Samy and Bengio, Yoshua},
  journal={Advances in Neural Information Processing Systems},
  volume={14},
  year={2001}
}

@inproceedings{aljundi2017expert,
  title={Expert gate: Lifelong learning with a network of experts},
  author={Aljundi, Rahaf and Chakravarty, Punarjay and Tuytelaars, Tinne},
  booktitle={Proceedings of the IEEE conference on computer vision and pattern recognition},
  pages={3366--3375},
  year={2017}
}

@article{fei2024scaling,
  title={Scaling diffusion transformers to 16 billion parameters},
  author={Fei, Zhengcong and Fan, Mingyuan and Yu, Changqian and Li, Debang and Huang, Junshi},
  journal={arXiv preprint arXiv:2407.11633},
  year={2024}
}

@article{dou2023loramoe,
  title={Loramoe: Revolutionizing mixture of experts for maintaining world knowledge in language model alignment},
  author={Dou, Shihan and Zhou, Enyu and Liu, Yan and Gao, Songyang and Zhao, Jun and Shen, Wei and Zhou, Yuhao and Xi, Zhiheng and Wang, Xiao and Fan, Xiaoran and others},
  journal={arXiv preprint arXiv:2312.09979},
  volume={4},
  number={7},
  year={2023}
}

@article{liu2023moelora,
  title={Moelora: An moe-based parameter efficient fine-tuning method for multi-task medical applications},
  author={Liu, Qidong and Wu, Xian and Zhao, Xiangyu and Zhu, Yuanshao and Xu, Derong and Tian, Feng and Zheng, Yefeng},
  journal={CoRR},
  year={2023}
}

@article{zhao2025each,
  title={Each Rank Could be an Expert: Single-Ranked Mixture of Experts LoRA for Multi-Task Learning},
  author={Zhao, Ziyu and Zhou, Yixiao and Zhu, Didi and Shen, Tao and Wang, Xuwu and Su, Jing and Kuang, Kun and Wei, Zhongyu and Wu, Fei and Cheng, Yu},
  journal={arXiv preprint arXiv:2501.15103},
  year={2025}
}

@inproceedings{ledig2017photo,
  title={Photo-realistic single image super-resolution using a generative adversarial network},
  author={Ledig, Christian and Theis, Lucas and Husz{\'a}r, Ferenc and Caballero, Jose and Cunningham, Andrew and Acosta, Alejandro and Aitken, Andrew and Tejani, Alykhan and Totz, Johannes and Wang, Zehan and others},
  booktitle={Proceedings of the IEEE conference on computer vision and pattern recognition},
  pages={4681--4690},
  year={2017}
}

@article{wang2023prolificdreamer,
  title={Prolificdreamer: High-fidelity and diverse text-to-3d generation with variational score distillation},
  author={Wang, Zhengyi and Lu, Cheng and Wang, Yikai and Bao, Fan and Li, Chongxuan and Su, Hang and Zhu, Jun},
  journal={Advances in Neural Information Processing Systems},
  volume={36},
  pages={8406--8441},
  year={2023}
}

@article{wang2024hero,
  title={Hero-SR: One-Step Diffusion for Super-Resolution with Human Perception Priors},
  author={Wang, Jiangang and Fan, Qingnan and Zhang, Qi and Liu, Haigen and Yu, Yuhang and Chen, Jinwei and Ren, Wenqi},
  journal={arXiv preprint arXiv:2412.07152},
  year={2024}
}

@inproceedings{mou2022metric,
  title={Metric learning based interactive modulation for real-world super-resolution},
  author={Mou, Chong and Wu, Yanze and Wang, Xintao and Dong, Chao and Zhang, Jian and Shan, Ying},
  booktitle={European Conference on Computer Vision},
  pages={723--740},
  year={2022},
  organization={Springer}
}

@article{chen2024topiq,
  title={Topiq: A top-down approach from semantics to distortions for image quality assessment},
  author={Chen, Chaofeng and Mo, Jiadi and Hou, Jingwen and Wu, Haoning and Liao, Liang and Sun, Wenxiu and Yan, Qiong and Lin, Weisi},
  journal={IEEE Transactions on Image Processing},
  year={2024},
  publisher={IEEE}
}

@inproceedings{golestaneh2022no,
  title={No-reference image quality assessment via transformers, relative ranking, and self-consistency},
  author={Golestaneh, S Alireza and Dadsetan, Saba and Kitani, Kris M},
  booktitle={Proceedings of the IEEE/CVF winter conference on applications of computer vision},
  pages={1220--1230},
  year={2022}
}

@inproceedings{chengdiff,
  title={Diff-MoE: Diffusion Transformer with Time-Aware and Space-Adaptive Experts},
  author={Cheng, Kun and He, Xiao and Yu, Lei and Tu, Zhijun and Zhu, Mingrui and Wang, Nannan and Gao, Xinbo and Hu, Jie},
  booktitle={Forty-second International Conference on Machine Learning}
}

@inproceedings{tu2023toward,
  title={Toward accurate post-training quantization for image super resolution},
  author={Tu, Zhijun and Hu, Jie and Chen, Hanting and Wang, Yunhe},
  booktitle={Proceedings of the IEEE/CVF Conference on Computer Vision and Pattern Recognition},
  pages={5856--5865},
  year={2023}
}

@article{tu2024ipt,
  title={IPT-V2: Efficient Image Processing Transformer using Hierarchical Attentions},
  author={Tu, Zhijun and Du, Kunpeng and Chen, Hanting and Wang, Hailing and Li, Wei and Hu, Jie and Wang, Yunhe},
  journal={arXiv preprint arXiv:2404.00633},
  year={2024}
}

@inproceedings{ding2025rass,
  title={RaSS: Improving Denoising Diffusion Samplers with Reinforced Active Sampling Scheduler},
  author={Ding, Xin and Yu, Lei and Li, Xin and Tu, Zhijun and Chen, Hanting and Hu, Jie and Chen, Zhibo},
  booktitle={Proceedings of the Computer Vision and Pattern Recognition Conference},
  pages={12923--12933},
  year={2025}
}
}

\appendix

\section{Degradation estimation module}
   
We propose a degradation estimation module that leverages the cross-modal alignment capability of the CLIP model. Specifically, we compute the cosine similarity between low-resolution (LR) images and predefined positive/negative textual descriptors of image quality (e.g., ``high-resolution, fine Details" vs. ``low-resolution, coarse artifacts"). The degradation severity of each input is then quantified as a normalized score derived from these similarity measures. For text prompt design, we adopt a principled strategy inspired by ~\cite{wang2022exploring,wang2024hero}, with detailed positive-negative text pairs provided in Table~\ref{tab:pos_neg_prompt}.
   
\begin{table}
  \centering
  \caption{Evaluation dimensions for image quality assessment and their associated positive and negative prompts. Lower image degradation corresponds to a smaller distance between the image and the positive prompt in CLIP embedding space.}
  \label{tab:pos_neg_prompt}
  \small
  \begin{tabular}{c|c|c}
     \toprule
     Evaluation dimension & Positive Prompts  & Negative Prompts \\
      \midrule
       Overall Quality & Good Image  & Bad Image  \\
       Blurriness & Sharp Image  & Blurry Image  \\
       Noise      & Noise-free image & Noisy Image \\
       Resolution & High-Resolution  & Low-Resolution \\
      Edge Clarity  & Sharp Edge  & Blurry Edge \\
      Clarity & Clear image & Vague Image \\
      Details & Fine Details & Coarse Details   \\
     \bottomrule
  \end{tabular} 
\end{table}

\begin{table}
  \centering
  \caption{We define two sets of degradation parameters for the experiments. The same dataset is degraded using the second-order degradation method proposed by Real-ESRGAN~\cite{wang2021real}, based on the degradation parameters outlined below, and the degraded data are fed into MoR-DASR. Subsequently, we analyzed the activation of MoR-DASR's experts under these two settings.}
  \label{tab:degradation}
  \small
  \begin{tabular}{c|c|c}
     \toprule
      Degradation parameter & Degradation 1 & Degradation 2 \\
      \midrule
      Noise range & [1,15]  & [1,30]  \\
      Poisson scale range & [0.05,1]  & [0.05,3]  \\
      Jpeg range      & [60,95] & [30,95] \\
      Second blur prob & 0.5  & 0.8 \\
      Noise range2 & [1,12]  & [1,25] \\
      Poisson scale range2      & [0.05,1] & [0.05,2.5] \\
      Jpeg range2      & [60,100] & [30,95] \\
      Blur kernel size2 & 11 & 21 \\
      Blur sigma2 & [0.2,1.0] & [0.2,1.5] \\
      Betag range2 & [0.5,2.0] & [0.5,4.0] \\
      Betap range2 & [1,1.5] & [1,2] \\
     \bottomrule
  \end{tabular} 
\end{table}

\begin{figure*}
    \centering
    \includegraphics[width=0.95\linewidth]{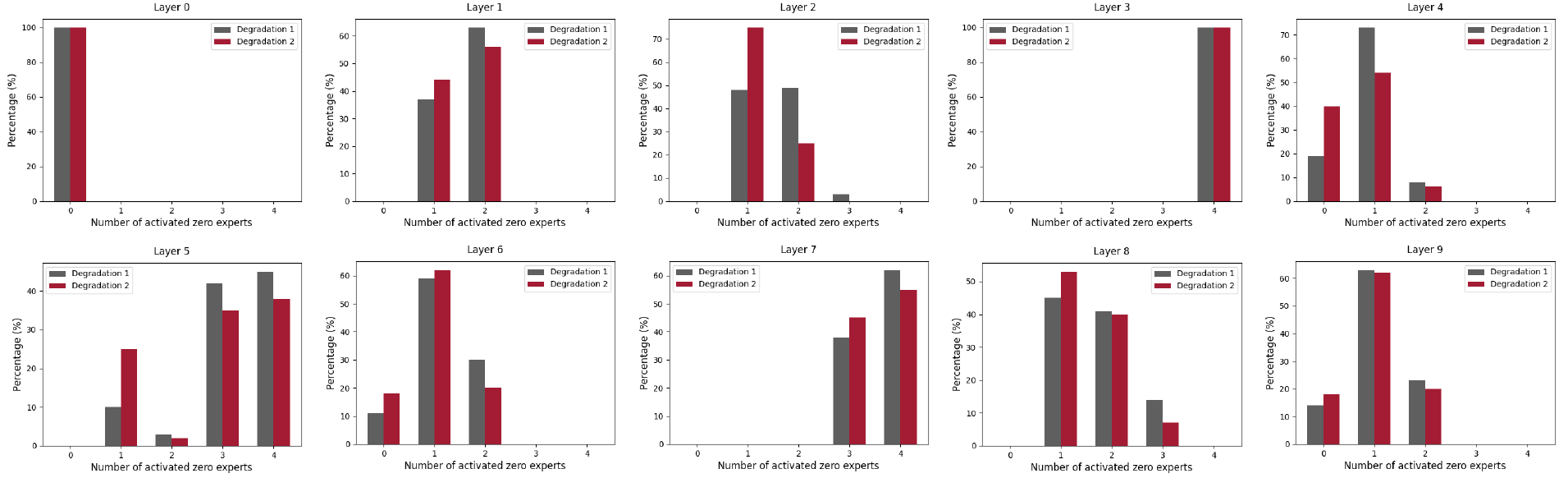}
    \caption{The visualization of activation counts for zero experts across 10 different MoR layers under two sets of degraded data is presented. The figure reveals significant variations in the number of activated zero experts at different positions within the layers, suggesting that the optimal rank required for each MoR layer differs. Additionally, data with more severe degradation tend to activate fewer zero experts, whereas data with milder degradation tend to activate more. }
    \label{fig:visual_expert}
 \end{figure*}

\begin{figure*}
    \centering
    \includegraphics[width=0.95\linewidth]{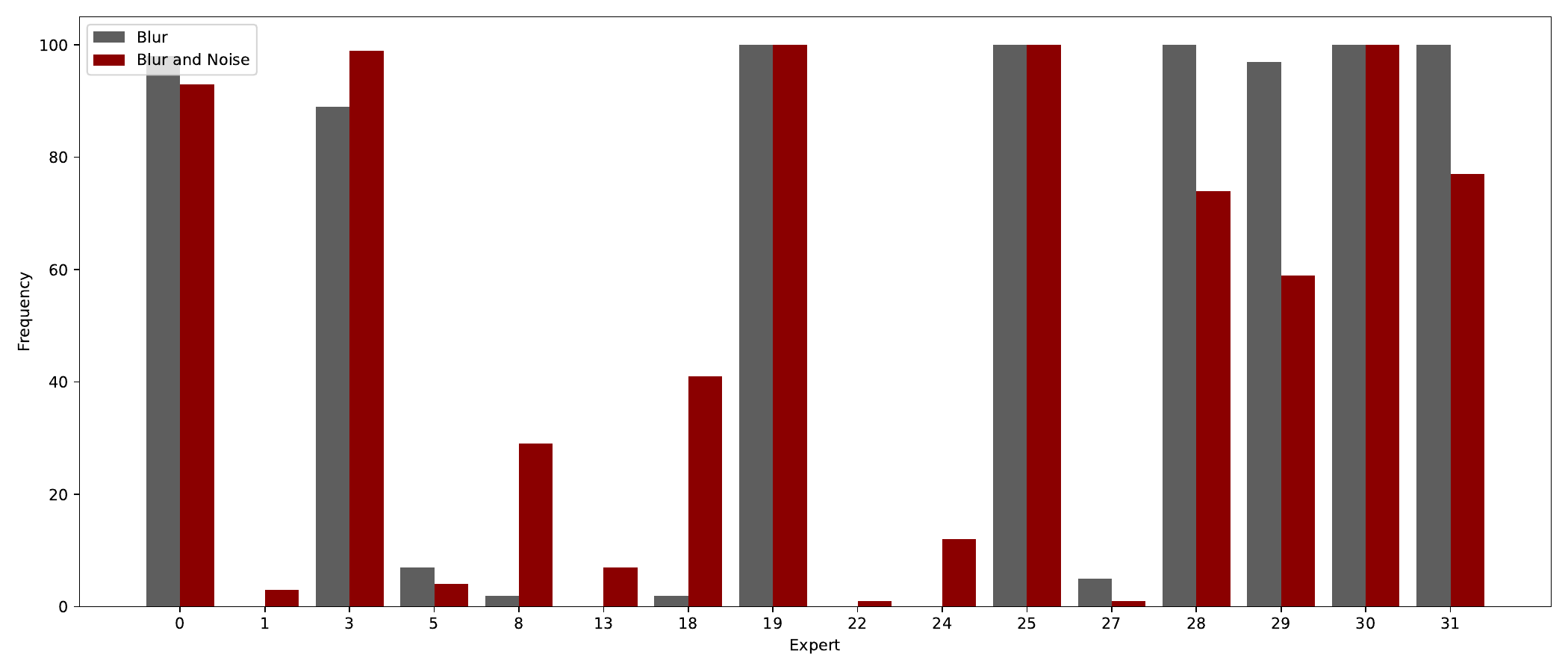}
    \caption{Visualization of expert activation patterns in a specific layer of MoR under various degradation conditions. }
    \label{fig:visual_expert-one_layer}
 \end{figure*}
 
\section{Expert Activation Analysis}

In this section, we analyze the routing mechanism of MoR-DASR through a visual and statistical lens. We first quantify expert activation patterns by evaluating the number of activated zero experts engaged across 100 images under two degradation severity regimes: the latter exhibiting higher degradation severity than the former (detailed degradation information is shown in Table~\ref{tab:degradation}). Figure~\ref{fig:visual_expert} visualizes expert activation across 10 MoR layers spanning network depth. Key observations include: 1. Degradation-Adaptive Activation. When the degree of image degradation increases, the number of activated experts will decrease, which is consistent with our intuition. 2. Layer-Specific Computation Divergence. Zero expert activation exhibits divergent patterns across layers, including two extreme cases: one is that all zero experts in that layer are activated, and the other is that no zero experts in that layer are activated. This phenomenon indicates that the capacity required by different layers of the network is indeed different, and our fine-grained MoR architecture effectively meets this demand by introducing zero experts.

\begin{table*}
    \centering
    \caption{Quantitative comparison with state of the arts multi-step Real-ISR methods. The best and second best results are highlighted in \textbf{bold} and \underline{underline}.}
    \label{tab:comparison_multi_step}
    \begin{tabular}{c|c|c|cccccccc}
        \toprule
         Datasets & Methods &Step  &PSNR$\uparrow$ &LPIPS $\downarrow$   &CLIPIQA$\uparrow$ & MUSIQ$\uparrow$   & MANIQA$\uparrow$ & TOPIQ$\uparrow$  & TRES$\uparrow$ \\
            \midrule
           \multirow{6}*{DIV2K-Val} 
            &StableSR &200 & 23.29  & \underline{0.312} & 0.675 &65.83 & 0.422  &0.598 &77.37 \\
            &DiffBIR &50 &23.64  &0.352 &0.670 &65.81 &\underline{0.475}  &0.634 &80.99 \\
            &PASD &20 & \underline{24.51} & 0.392 &0.551 & 59.99 &0.399  &0.466 &62.21 \\
            &SeeSR &50 & 23.68  & 0.319 &\textbf{0.693} & \textbf{68.68} & \textbf{0.504}  & \textbf{0.686} & \textbf{85.80} \\
            &ResShift &15  &\textbf{24.72} &  0.34 &  0.594  &   60.89   &  0.399  &0.525 &73.35  \\
            &MoR-DASR &1 &24.01 &\textbf{0.289} &\underline{0.681} &\underline{68.09} &\underline{0.475}  &\underline{0.663} &\underline{84.14} \\ 
            \midrule
            \multirow{6}*{RealSR}
            &StableSR &200 & 25.63 & 0.307  & 0.528 &61.11 & 0.366  &0.575 & 74.26 \\
            &DiffBIR &50 &24.24 &0.347 &0.654 &64.25 &0.485  &0.605  &78.99 \\
            &PASD &20 & \textbf{26.67} & 0.344 &0.519 & 62.92 &0.404 &0.523 &66.99 \\
            &SeeSR &50 & 25.24  & \underline{0.301} &\underline{0.669} & \textbf{69.82} & \textbf{0.540} &\textbf{0.688} & \textbf{88.60} \\
            &ResShift &15 & \underline{26.34}  &  0.346 &  0.542  &  56.06   &  0.375 &0.522 &73.76 \\
            &MoR-DASR &1 &25.32 & \textbf{0.291} &\textbf{0.691} &\underline{69.78} & \underline{0.512}  & \underline{0.662} & \underline{84.97} \\   
            \midrule
            \multirow{6}*{DRealSR}
            &StableSR &200 & 28.24 & \underline{0.315} & 0.606 &57.42 & 0.369 &0.532 &66.73\\
            &DiffBIR &50 &25.93 &0.452 &0.686 &63.47 &0.492 &0.615 &77.87\\
            &PASD &20 & \textbf{29.06} & \underline{0.315} &0.538 & 55.33 &0.387 &0.518 &64.16 \\
            &SeeSR &50 & 28.09 & 0.319 & \underline{0.691} & \underline{65.08} & \textbf{0.513} &\textbf{0.657} & \textbf{84.62} \\
            &ResShift &15 & 28.27  &  0.401 &  0.529  &   50.14   &  0.328 &0.476 &64.29  \\
            &MoR-DASR &1 &\underline{28.37} &\textbf{0.307} &\textbf{0.717} & \textbf{65.94} & \underline{0.509} & \underline{0.652} & \underline{81.78} \\ 
      \bottomrule
    \end{tabular} 
  \end{table*}

In addition, we conduct further experiments to analyze the expert activation patterns in a specific layer of the network. Specifically, we select 100 images as the test set and create two groups: one with added noise degradation and the other with additional blur degradation on top of the noise. We randomly select one layer from the network to monitor expert activations. As shown in Figure~\ref{fig:visual_expert-one_layer}, under noise-only degradation, the model heavily activates the null experts (experts 28–31), while relying primarily on a small subset of real experts (experts 0, 3, 19, and 25) for restoration. However, when additional blur degradation is introduced, the activation frequency of null experts decreases sharply, and real experts such as 1, 13, 22, and 24 begin to be activated. In particular, experts 8 and 18 exhibit a notable increase in activation frequency, suggesting that they may have learned representations related to handling blur. These results demonstrate that our model is capable of adaptively activating different expert combinations in response to varying degradation types and levels.

\section{Extended Experimental Results}

\subsection{Comparison with multi-step Real-ISR methods}
In this section, we compare MoR-DASR with state-of-the-art multi-step real super-resolution methods, including DiffBIR~\cite{lin2023diffbir}, StableSR~\cite{wang2024exploiting}, PASD~\cite{yang2023pixel}, ReShift~\cite{yue2024resshift} and SeeSR~\cite{wu2024seesr}. As shown in Table~\ref{tab:comparison_multi_step}, our method achieves performance comparable to SeeSR on no-reference metrics such as CLIPIQA and MUSIQ, and ranks second only to SeeSR on MANIQA and TRES, outperforming all other competing methods. Moreover, MoR-DASR surpasses SeeSR across all reference-based metrics, including PSNR and LPIPS, further validating the effectiveness of our approach. In terms of efficiency, as reported in Table~\ref{tab:runtime1}, our method generates high-resolution images using only a single sampling step. Compared to SeeSR, the inference speed is increased by 40$\times$, demonstrating a substantial advantage in computational efficiency for real-world deployment.

\subsection{Comparison with GAN-based Real-ISR Methods}
In addition to benchmarking MoR-DASR against diffusion-based Real-ISR methods, we further evaluate its performance against state-of-the-art GAN-based approaches, including BSRGAN~\cite{zhang2021designing}, Real-ESRGAN~\cite{wang2021real}, LDL~\cite{liang2022details}, and FeMASR~\cite{chen2022femasr}. As shown in Table~\ref{tab:GAN-based}, while GAN-based methods achieve marginally higher scores on full-reference metrics such as PSNR and SSIM, we posit that this discrepancy arises from MoR-DASR's ability to synthesize richer perceptual details, which may introduce deviations from ground-truth content. Crucially, MoR-DASR maintains competitive performance on full-reference metrics while significantly outperforming GAN-based methods on non-reference metrics. These non-reference metrics prioritize perceptual quality and are closely aligned with human perceptual evaluations. It demonstrates that our framework effectively balances fidelity to the LR image and perceptual quality, prioritizing human visual preferences over strict pixel-level conformity.

\begin{table*}
    \centering
    \caption{Quantitative comparison with state of the arts GAN-based Real-ISR methods. The best and second best results are highlighted in \textbf{bold} and \underline{underline}.}
    \label{tab:GAN-based}
    \begin{tabular}{@{}C{2.1cm}@{}|@{}C{2.2cm}@{}|cccccccc}
        \toprule
         Datasets & Method  &PSNR$\uparrow$ &SSIM$\uparrow$ &LPIPS $\downarrow$   &CLIPIQA$\uparrow$ & MUSIQ$\uparrow$   & MANIQA$\uparrow$ & TOPIQ$\uparrow$  & TRES$\uparrow$ \\
            \midrule
           \multirow{5}*{DIV2K-Val} 
            &BSRGAN  & \textbf{24.58} &0.627 & 0.335  &0.524  & \underline{61.19} &0.356  & \underline{0.546} & \underline{74.03} \\
            &Real-ESRGAN  & \underline{24.29} & \textbf{0.637} & \underline{0.311}  &0.527 &61.06 & \underline{0.382} &0.530 &70.13 \\
            &LDL &23.83  &\underline{0.634} &0.326  &0.518 &60.04  &0.375  &0.514 &68.25 \\
            &FeMASR  & 23.06 & 0.589 & 0.313  & \underline{0.599} & 60.83 &0.346  & 0.523 & 70.73 \\
            &MoR-DASR &24.01 &0.606 &\textbf{0.289} &\textbf{0.681} &\textbf{68.09} &\textbf{0.475}  &\textbf{0.663} &\textbf{84.14} \\ 
            \midrule
            \multirow{5}*{RealSR}
            &BSRGAN  & \textbf{26.38} & \textbf{0.765} & \textbf{0.267}    & \underline{63.28} & \underline{0.376} & \underline{0.562} & \underline{0.551} & \underline{75.70} \\
            &Real-ESRGAN  &25.06 &0.736 &0.294  &0.449 &59.06 &0.373 &0.503 &67.21 \\
            &LDL &25.28  &0.757 &0.277  &0.448 &60.82  &0.342 &0.512 &68.60 \\
            &FeMASR  & \underline{25.69} & \underline{0.761} & \underline{0.271} & \underline{0.541} & 60.37 &0.361  & 0.515 & 67.68 \\
            &MoR-DASR &25.32 & 0.728 & 0.291 &\textbf{0.691} &\textbf{69.78}   & \textbf{0.651} & \textbf{0.662} & \textbf{84.97} \\   
            \midrule
            \multirow{5}*{DRealSR}
            &BSRGAN  & \textbf{28.70} &0.803 & 0.286 &0.492 & \underline{57.16} &0.343  &\underline{0.506} &\underline{66.76} \\
            &Real-ESRGAN  &26.87 &0.757 &0.316  &0.442 &53.70 & 0.344  &0.467 &59.35 \\
            &LDL  &28.21  &\textbf{0.813} &\textbf{0.281}   &0.431 &53.85  & \underline{0.345} &0.490 &58.82 \\
            &FeMASR  & \underline{28.61} & \underline{0.804} & \underline{0.282}  & \underline{0.546} & 54.28 & 0.332  & 0.462 & 58.79\\
            &MoR-DASR &28.37 &0.776 &0.307 &\textbf{0.717} & \textbf{65.94} & \textbf{0.509}   & \textbf{0.652} & \textbf{81.78} \\ 
      \bottomrule
    \end{tabular} 
  \end{table*}
  
\begin{table*}
    \centering
    \caption{Complexity comparison among different SR methods. All methods are tested on the $\times$4 (128$\rightarrow$512) SR tasks, and the inference time is measured on an V100 GPU.}
    \label{tab:runtime1}
    \begin{tabular}{c|c c c c c c c c c c}
    \Xhline{0.8pt}
    Method & StableSR  &DiffBIR &SeeSR & ResShift  & SinSR & AddSR &OSEDiff &S3Diff & MoR-DASR \\
    \Xhline{0.4pt}
    NFE & 200 & 50 &50 &15 &1 &1 &1 &1 &1  \\
    Inference time (s) &15.32  & 11.71   & 8.23 & 1.41  &0.181  & 0.455 & 0.178 & 0.521 &0.186  \\
    Trainable Param (M) &150.0 &380.0 &749.9 &118.6 &118.6 &749.9&8.5 & 34.5 &54.0\\
    \Xhline{0.4pt}
    \end{tabular} 
 \end{table*}

\subsection{Ablation Study}

\textbf{Effectiveness of degradarion estimation module. }
To validate the effectiveness of our degradation-aware estimation module in guiding expert routing, we present ablation studies in Table~\ref{tab:ablation_deg}. We evaluate two variants: (1) a baseline using low-resolution (LR) image encoding features for routing, and (2) using the degradation estimation module proposed in MM-RealSR~\cite{mou2022metric} to guide routing, which primarily estimates the noise and blur levels of the image. The results shows that our degradation-aware module achieves superior performance, outperforming both alternatives. The degradation estimation module proposed in MM-RealSR underperformed in our experiments. We attribute this to its reliance on noise and blur metrics alone, which are insufficient for real-world degradation.

\textbf{Impact of zero experts.}
In our MoR framework, zero experts are introduced to enable dynamic computational resource allocation based on input samples. To assess their impact, we conduct an ablation study by progressively substituting real experts with zero experts while maintaining a fixed total rank of 32 for routed experts (8 activated per sample). As shown in Table~\ref{tab:zero_expert}, while zero experts initially enhance performance by dynamic computations allocation, excessive zero expert degrades results due to underutilization of critical model capacity. Consequently, we set the number of zero experts in our proposed MoR to 4.

\textbf{Impact of prompt design in degradation estimation module.}
Our predefined prompts go beyond simple positive and negative text pairs related to degradation types such as noise and blur. They are also designed from the perspective of human perception of image quality, incorporating aspects such as detail, edge sharpness, overall visual quality, and resolution (as shown in Table~\ref{tab:pos_neg_prompt}). Regardless of the specific degradation type, any form of degradation inevitably lowers the visual quality, increasing the distance between the degraded image and the positive text embedding used to evaluate image quality. Consequently, the degradation score rises, allowing our model to detect degradation more effectively. This design makes our degradation estimation module inherently robust to unseen degradation. We further validate the effectiveness of prompts related to perceptual image quality through ablation studies, as shown in Table~\ref{tab:prompt design}. When the prompts contain only degradation-related cues (noise and blur), the model performance declines. In contrast, incorporating perceptual quality cues—such as detail, resolution, and overall visual quality—leads to a noticeable performance improvement.



\textbf{Impact of MoR ranks on model performance.} We explore the impact of different MoR ranks on model performance, and the results are shown in Table D. As shown in Table~\ref{tab:MoR_ranks}, increasing the rank from 20 to 40 leads to noticeable improvements in various evaluation metrics on the real-world dataset. However, when the rank is further increased from 40 to 60, the model's performance plateaus, indicating that simply increasing the ranks beyond a certain point does not yield additional benefits.

\textbf{Impact of top-k on router.} We have referenced existing MoE methods~\cite{dai2024deepseekmoe,chengdiff} and activated 1/4 of the total experts. Additionally, we conducted an ablation study on the number of activated experts while keeping the total number of experts fixed, as shown in Table~\ref{tab:topk}. When fewer experts are activated, the model's expressive capacity may be insufficient, leading to a slight decrease in performance. Conversely, when half of the experts are activated, the specialization of the experts is reduced, and the gradient updates for each expert become diluted during training, resulting in a decline in performance. Overall, activating 8 experts (ranks) out of the 32 routing experts (ranks) yielded the best performance.

\textbf{Impact of loss functions.} We clarify that our baseline method is OSEDiff, and our proposed approach primarily focuses on the design of the MoR module. Regarding the loss design, we largely adopt the configuration from OSEDiff. For the additional GAN loss and degradation-aware load balancing loss (denote as deg-aware loss), we conduct ablation experiments as shown in Table~\ref{tab:loss_ablation}. The results demonstrate that the GAN loss effectively enhances image quality while maintaining visual fidelity, whereas our proposed load balancing loss enables the model to adaptively activate experts based on the difficulty of the samples, thereby improving overall performance.

\begin{table}
    \centering
    \caption{Ablation study of degradation estimation module. We integrate three distinct inputs into the routing mechanism: (1) LR image features, (2) noise and blur scores from MM-RealSR~\cite{mou2022metric}, and (3) degradation-aware representations generated by our proposed module.}
    \label{tab:ablation_deg}
    \begin{tabular}{c|ccc}
    \Xhline{0.8pt}
    Router input  & CLIPIQA $\uparrow$  & MANIOA $\uparrow$& TRES$\uparrow$ \\
    \Xhline{0.4pt}
    LR feature  & 0.697 & 0.495 & 81.21   \\
    MM-RealSR  & 0.695 & 0.490 & 80.75  \\
    Ours & \textbf{0.717} & \textbf{0.509} & \textbf{81.78}  \\
    \Xhline{0.4pt}
    \end{tabular} 
 \end{table}

 \begin{table}
    \centering
    \caption{Ablation study of varying numbers of zero experts in the MoR module.}
    \label{tab:zero_expert}
    \begin{tabular}{c|ccc}
    \Xhline{0.8pt}
    Zero experts  & CLIPIQA $\uparrow$  & MANIOA $\uparrow$& TRES$\uparrow$\\
    \Xhline{0.4pt}
    0  & 0.704 & 0.491 & 80.32   \\
    4  & \textbf{0.717} & \textbf{0.509} & \textbf{81.78}  \\
    8  & 0.695 & 0.476 & 78.48  \\
    \Xhline{0.4pt}
    \end{tabular} 
 \end{table}

 \begin{table}
    \centering
    \caption{Impact of prompt design on model performance.}
    \label{tab:prompt design}
    \begin{tabular}{c|ccc}
    \Xhline{0.8pt}
    Prompt  & CLIPIQA $\uparrow$  & MANIOA $\uparrow$& TRES$\uparrow$\\
    \Xhline{0.4pt}
    blur and noise   &0.694 & 0.489 & 80.55   \\
    + image quality &\textbf{0.717}  &\textbf{0.509}  &\textbf{81.78}  \\
    \Xhline{0.4pt}
    \end{tabular} 
 \end{table}

 \begin{table}
    \centering
    \caption{Quantitative results of different MoR ranks on DRealSR dataset.}
    \label{tab:MoR_ranks}
    \begin{tabular}{c|ccc}
    \Xhline{0.8pt}
    Zero experts  & CLIPIQA $\uparrow$  & MANIOA $\uparrow$& TRES$\uparrow$\\
    \Xhline{0.4pt}
    20 & 0.687 & 0.03 & 80.44  \\
    40  &\textbf{0.717}  & \textbf{0.509} & 81.78   \\
    60 & 0.703 & 0.506 & \textbf{82.25}  \\
    \Xhline{0.4pt}
    \end{tabular} 
 \end{table}

  \begin{table}
    \centering
    \caption{Quantitative results of different top-k settings of router on DRealSR dataset.}
    \label{tab:topk}
    \begin{tabular}{c|ccc}
    \Xhline{0.8pt}
    Activated experts  & CLIPIQA $\uparrow$  & MANIOA $\uparrow$& TRES$\uparrow$\\
    \Xhline{0.4pt}
    4 & 0.696 & 0.482 & 79.83  \\
    8   &\textbf{0.717}  & \textbf{0.509} & \textbf{81.78}   \\
    16 & 0.664 & 0.451 & 78.93  \\
    \Xhline{0.4pt}
    \end{tabular} 
 \end{table}

\begin{table}
    \centering
    \caption{Loss ablation on DRealSR dataset.}
    \label{tab:loss_ablation}
    \begin{tabular}{c|ccc}
    \Xhline{0.8pt}
    Methods  & CLIPIQA $\uparrow$  & MANIOA $\uparrow$& TRES$\uparrow$\\
    \Xhline{0.4pt}
    w/o GAN loss & 0.664 & 0.451 & 74.83  \\
    w/o deg-aware loss & 0.699 & 0.479 & 79.41 \\
    Ours  &\textbf{0.717}  & \textbf{0.509} & \textbf{81.78}   \\
    \Xhline{0.4pt}
    \end{tabular} 
 \end{table}

\section{More Visual Comparisons}

\begin{figure*}
    \centering
    \includegraphics[width=0.95\linewidth]{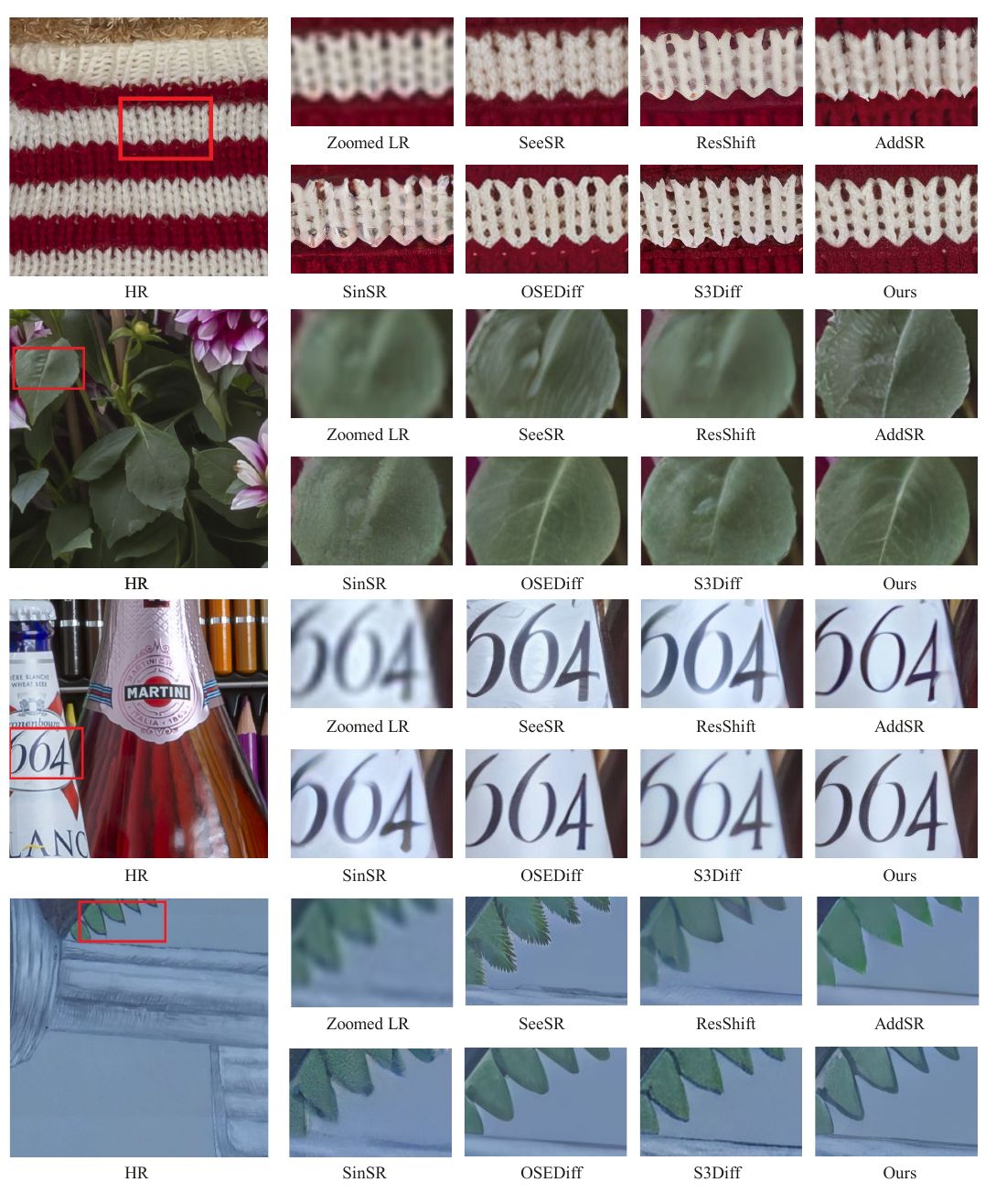}
    \caption{Visual comparisons of different Real-ISR methods. Please zoom in for a better view.}
    \label{fig:appendix_results1}
 \end{figure*}

\begin{figure*}
  \centering
  \includegraphics[width=0.95\linewidth]{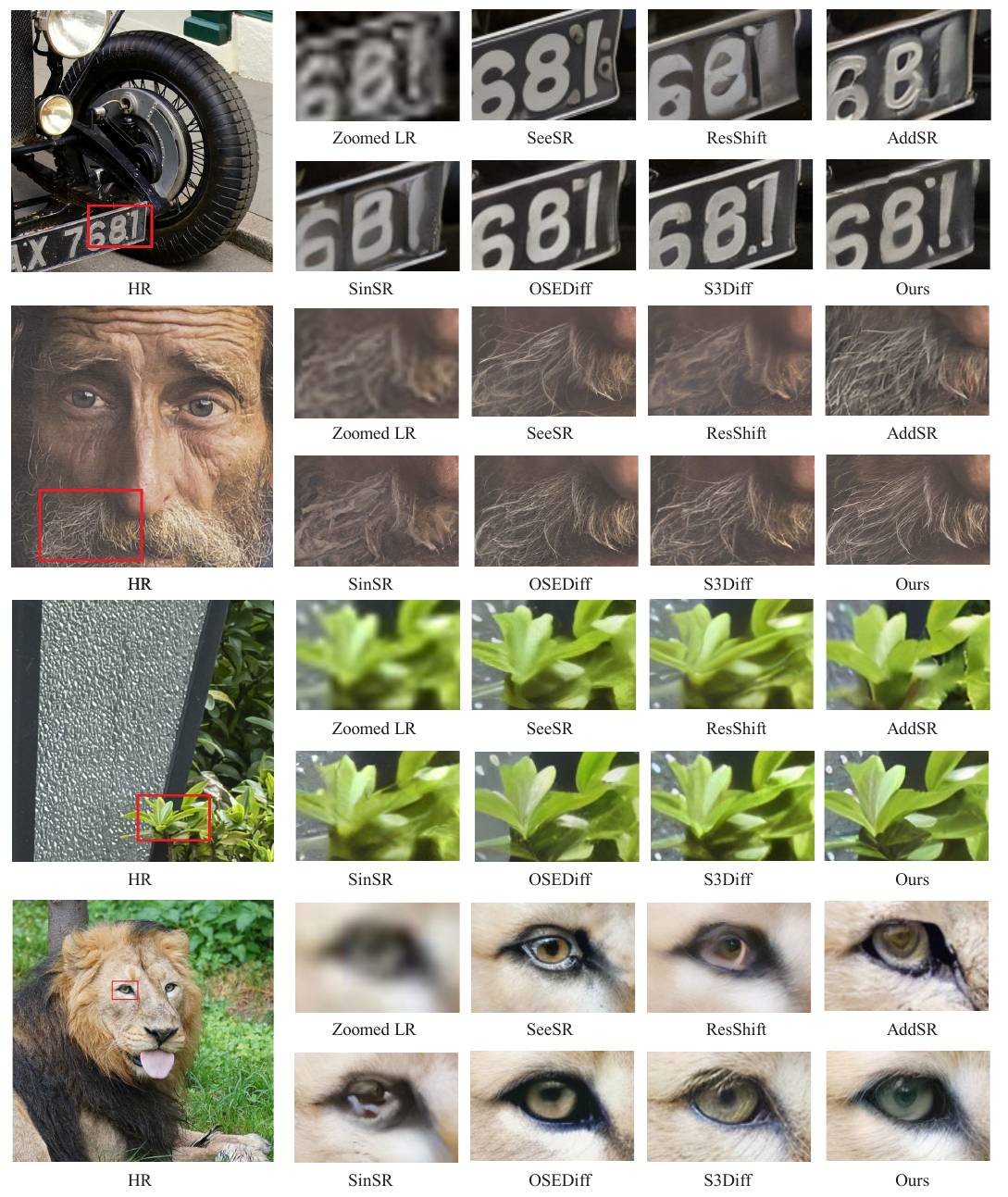}
  \caption{Visual comparisons of different Real-ISR methods. Please zoom in for a better view.}
  \label{fig:appendix_results2}
\end{figure*}

Figures~\ref{fig:appendix_results1}~\ref{fig:appendix_results2} present an extended qualitative comparison with state-of-the-art diffusion-based Real ISR methods. These visualized empirical results highlight the strong restoration capability of MoR-DASR, which generates high-definition textures while preserving structural integrity and minimizing the occurrence of illusory artifacts.

\end{document}